# Supervised learning of a regression model based on latent process. Application to the estimation of fuel cell life time

Raïssa Onanena[(1)], Faicel Chamroukhi[(1)], Latifa Oukhellou[(1)(2)], Denis Candusso[(1)(4)], Patrice Aknin[(1)], Daniel Hissel[(3)(4)]

[(1)] INRETS-LTN, 2 av de la butte verte, 93166 Noisy le Grand Cedex, France
[(2)] CERTES-Université Paris 12, 61, av du Gal de Gaulle, 94100 Créteil, France
[(3)] FEMTO-ST UMR CNRS 6174, Université de Franche-Comté, 90010 Belfort Cedex, France
[(4)] FCLAB, Rue Ernest Thierry-Mieg, 90010 Belfort Cedex, France
{onanena, chamroukhi, oukhellou, candusso, aknin}@inrets.fr, daniel.hissel@univ-fcomte.fr

## Abstract

*This paper describes a pattern recognition approach aiming to estimate fuel cell duration time from electrochemical impedance spectroscopy measurements. It consists in first extracting features from both real and imaginary parts of the impedance spectrum. A parametric model is considered in the case of the real part, whereas regression model with latent variables is used in the latter case. Then, a linear regression model using different subsets of extracted features is used for the estimation of fuel cell time duration. The performances of the proposed approach are evaluated on experimental data set to show its feasibility. This could lead to interesting perspectives for predictive maintenance policy of fuel cell.*

## 1. Introduction

Fuel cells (FCs) are generally considered as promising and environmentally friendly energy-conversion solutions for the future. Among the different types of fuel cells, the polymer electrolyte fuel cells (PEFC) are prime candidates for applications in transport [1]. Indeed, they can offer high fuel economy, through higher efficiency and sustainably lower $CO_2$ emissions. However, considerable challenges still remain for the widespread marketing of FC generators in terms of durability, reliability and performance improvements.

Despite the apparent reliability of FCs –due to the absence of any moving part–, the stack itself is prone to material degradation which is strongly affected by the operating conditions (i.e., the amount of reactant gas flows versus the load current demand, operating temperature, mechanical constraints on the membrane electrode assemblies etc.) and some physical degradation causes (e.g. poisoning of the catalyst sites, loss of proton conductivity in the membrane, corrosion of plates,…) [2][3]. Therefore, there is a need for automatic diagnosis schemes that allow evaluating the state-of-health of the FC stack.

Various diagnosis approaches for FC stacks and systems have been developed. These include model-based approach [5][6] and gray or black-box model approaches using fuzzy logic [7], neural networks [8], or non-parametric identification by Markov parameters [9]. A recent FC stack diagnosis approach based on a fuzzy clustering has been proposed in [10]. All these contributions try to identify the working state of the FC. The last ones use particular measurements for the information retrieval: the electrochemical impedance spectrometry (EIS).

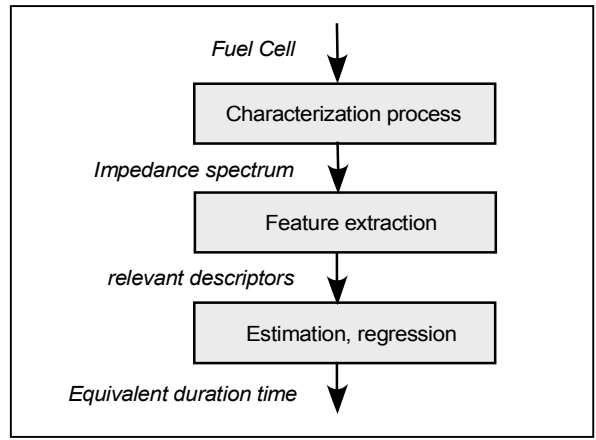

Figure 1: FC time duration estimation on the basis of pattern recognition approach

This paper presents a pattern recognition based approach aiming to estimate the duration of the fuel cell's operating from EIS measurement. For the FC technology, the knowledge of duration time (and dually the expectation of remaining working time) is essential for the definition of predictive maintenance strategies. The approach that we investigate can be summarized as shown in Figure 1. From the impedance spectrum, a feature extraction is first performed to automatically generate descriptors. Then, linear or non linear regressions between a subset of those descriptors and the considered output (duration time) are achieved.

The choice of representation space is essential, especially because the available training data set is sparse. In this context, a particular attention has to be made to use only relevant variables as inputs of the regression model, otherwise the well known phenomenon of curse of dimensionality could appear inevitably [11]. Therefore, part of the difficulty of feature extraction arises from the fact that the impedance spectrum has to be automatically partitioned into different segments that correspond to different behaviours of the FC. Rather than using a global fitting of the measurement curve, the partitioned parametrization, despite its complexity seems to be more relevant in our application. Furthermore, a selection of subset of meaningful variables from the original ones is carried out to keep only descriptors related to the FC ageing in the final regression step.

The paper is organized as follows. Section 2 describes the ageing tests and highlights the link between the FC ageing and the EIS measurement. Section 3 provides details on the feature extraction methods used to summarize real and imaginary parts of the FC impedance. Section 4 explains the different solutions for duration time estimation based on linear regression. Experimental results are reported into this section. Section 5 concludes the paper with some perspectives.

## 2. Impedance spectroscopy as a durability indicator

### 2.1. Ageing tests description

The durability tests were performed on two identical small power PEFC three-cell stacks of about 100W during 1000 hours. The testing conditions varied from one stack to another. The stack (noted as FC1) is operated under nominal and stationary conditions. The load current is constant and equal to 50A. Moreover, FC1 was operated in an open mode (i.e. atmospheric pressure): both anode and cathode flows were controlled by flow regulators placed upstream from the stack. In the second test (noted as FC2), the FC is operated under dynamical load current based on a real transportation mission profile (maximum current of 70A was reached for an average of 12.5A). Details on how the current solicitation was defined can be found in [2].

During both ageing tests, the stacks were characterised regularly (approx. once the week) through polarisation curves and impedance spectroscopy [2]. In this paper, the ageing process will be analysed with the information extracted from EIS spectra only.

### 2.2. Electrochemical impedance spectroscopy

Electrochemical impedance spectroscopy is a method that is commonly used by electrochemists in order to obtain a better understanding of electrochemical device behaviour [3]. It is a powerful technique that allows characterising the stack in dynamic conditions. The dynamical FC behaviour is carried out considering a static operating point (in our case, 35A) and a small sinusoidal alternating part around it (an amplitude of 1A and a frequency range from 10mHz to 30kHz). The real and imaginary parts of the FC impedance are calculated from the measured current and voltage alternating components. During the measurements, the stack remains at its nominal operating conditions (in terms of temperature, stoichiometry rates, and hygrometry).

Figure 2 presents an impedance plot (imaginary part function of real part) that was measured during the ageing tests on FC1. As it can be seen on this figure, the impedance spectrum could be divided into three parts where each one of them corresponds to a specific behaviour of the stack:
- an inductive part which is present in high frequencies (4kHz < $f$),
- a first capacitive arc (130Hz < $f$ < 4kHz),
- a second capacitive arc ($f$ < 130Hz).

Obviously, the delimiting frequency values are not precisely known.

### 2.3. Effect of FC ageing on impedance spectrum

The impedance plane is usually used to highlight the dynamic behaviour of the FC stack [4]. In this paper, we choose to work with the real and imaginary parts of the spectrum versus frequency in an explicit way. Thus, we are able to use the additional information laying into the frequency variable.

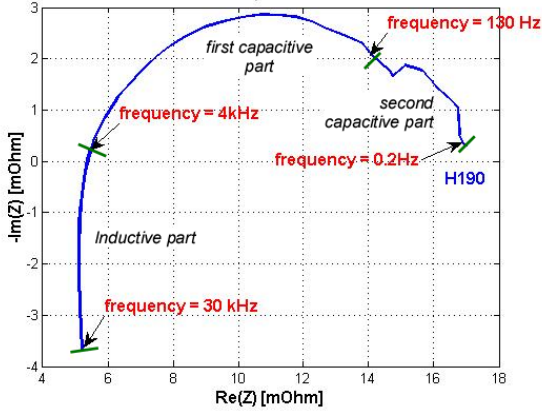

Figure 2: Evolution of the impedance spectrum for the FC1 in the impedance plane

The two sets (obtained for different operating times) of real and imaginary parts obtained for FC1 are presented on Figure 3 versus frequency. It can be noticed that an inherent link can be established between the ageing phenomenon and the evolution of these diagrams. Hence, the idea which consists to describe these plots with a reduced number of variables can be used to estimate the ageing time of the FC stack.

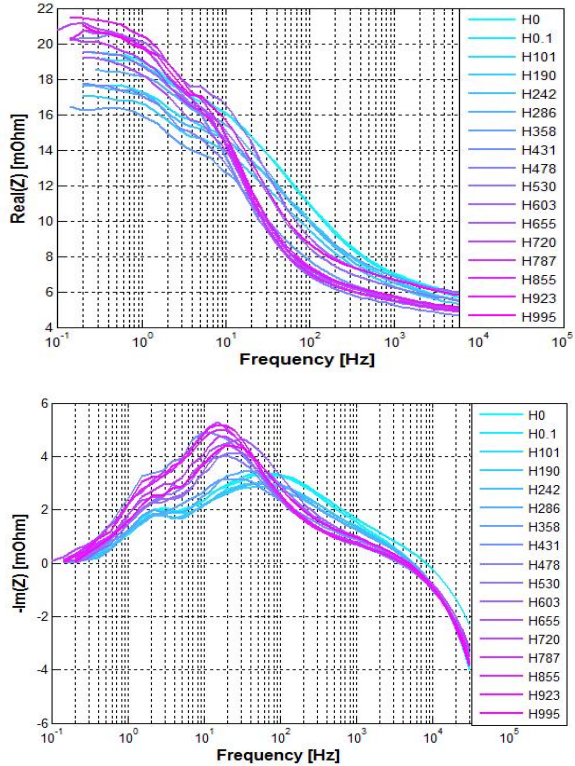

Figure 3: Evolution of the real and imaginary parts of impedance spectrum versus frequency for FC1 from the initial state (H=0) to the final state (H=1000)

## 3. Feature extraction

The spectrum data dimension is about 50 and the number of observations of the fuel cell spectrum during its ageing is about 20. So a feature extraction task is greatly recommended to reduce the dimension of the input space, summarize efficiently the measurements, and avoid the curse of dimensionality [11] [13].

The next section presents the methods used for feature extraction from the real and the imaginary impedance spectrum measurements.

### 3.1. Parametrization of the real part of the impedance

The real part of the impedance as a function of the frequency does not present a particular difficulty since it can be approximated by an external model of 4 parameters related to an extended "logsig" function and defined as follows:

$$\text{Re}(\log(f)) = \frac{a_1}{1+e^{-a_2(\log(f)-a_3)}} + a_4 \quad (1)$$

The model coefficients are determined by minimizing a cost function based on a mean square error with the help of simplex method [12]. Figure 4 shows the good behavior of this external model.

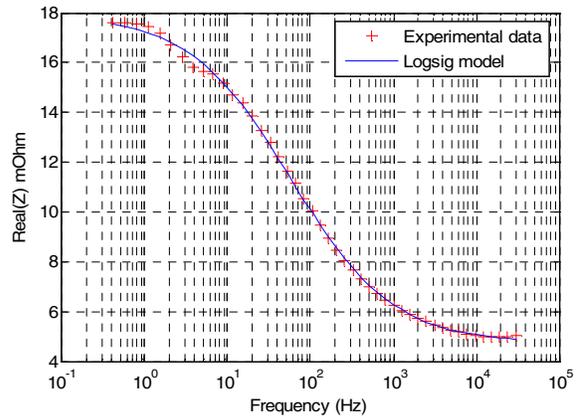

Figure 4: Example of real part of the impedance spectrum and its approximation by the external logsig model

The imaginary part of the spectrum is more informative and more complex than the real one. Particularly, the three domains mentioned in section 2.2 are perceptible (cf. Figure 3b). The following section introduces the proposed approach used for feature extraction from the imaginary part of spectrum.

## 3.2. Regression model with hidden logistic process

The feature extraction approach we use consists of a specific regression model incorporating a discrete hidden logistic process [14]. The estimated model parameters can be directly used as the feature vector for each measurement. This model is adapted for measurements including smooth or abrupt transitions in regimes.

Let $x = (x_1,...,x_n)$ be the $n$ points of the imaginary part of an impedance spectrum where $x_i$ is observed for the frequency value $f_i$. In the following, $f_i$ will denoted the logarithm of the frequency ($f_i \sim \log f_i$) that is more convenient for spectrum representation. This specific regression model assumes that the measurement incorporates $K$ polynomial regimes where the switching from one regime to another is automatically controlled by a latent discrete variable $z_i$ which takes its values in the set $\{1,...,K\}$. This latent variable represents the class label of the polynomial regression model generating $x_i$. Thus, the sample is assumed to be generated by the following regression model:

$$\forall i = 1...n, \quad x_i = \beta_{z_i}^T r_i + \sigma_{z_i} \varepsilon_i, \quad (2)$$

where the sequence of the latent variables $z=(z_1,...,z_n)$ is a logistic process. It allows the switching from one regression model to another in $K$ models.

This process assumes that the variables $z_i$, given the frequencies $(f_1,...,f_n)$, are generated independently according to the multinomial distribution $M(1,\pi_{i1}(w),...,\pi_{iK}(w))$, where

$$\pi_{ik}(w) = p(z_i = k; w) = \frac{\exp(w_{k0} + w_{k1}f_i)}{\sum_{j=1}^{K} \exp(w_{j0} + w_{j1}f_i)}, \quad (3)$$

is the logistic transformation of a linear function of the frequency $f_i$ and $w = (w_{10}, w_{11},..., w_{k0}, w_{k1},..., w_{K0}, w_{K1})^T$ is the parameter vector of the logistic process. The relevance of the logistic transformation in terms of flexibility of transition has been well detailed in [14].

## 3.3. Parameter estimation

From the model given by Eq. (2) it can be proven that the variable $x_i$ is distributed according to the normal mixture density [20]:

$$p(x_i; \theta) = \sum_{k=1}^{K} \pi_{ik}(w) \, \phi(x_i; \beta_k^T r_i, \sigma_k^2), \quad (4)$$

where $\phi(.; \mu, \sigma^2)$ denotes a monodimensional normal density with mean $\mu$ and variance $\sigma^2$ and $\theta$ the parameter vector to be estimated,

$\theta = (w, \beta_1,...,\beta_K, \sigma^2_1,...,\sigma^2_K)$

Assuming that, given $(f_1,...,f_n)$, the $x_i$ are independent, the log-likelihood of $\theta$ can be written as:

$$L(\theta; x_1,...,x_n) = \log \prod_{i=1}^{n} p(x_i; \theta) \\
= \sum_{i=1}^{n} \log \sum_{k=1}^{K} \pi_{ik}(w)\phi(x_i; \beta_k^T r_i, \sigma_k^2) \quad (5)$$

This likelihood cannot be directly maximized, then we use a dedicated Expectation Maximization (EM) algorithm [15][17] to perform the maximization. The parameters of the hidden logistic process, in the inner loop of the EM algorithm, are estimated using a multi-class Iterative Re-weighted Least-Squares (IRLS) algorithm [16].

## 3.4. Measurement approximation

In addition to perform feature extraction, this regression model can be used to approximate and segment the impedance spectrum. The approximated measurement at sample $i$ is given by the expectation:

$$E(x_i; \hat{\theta}) = \int_{-\infty}^{+\infty} x_i p(x_i; \hat{\theta}) dx_i = \sum_{k=1}^{K} \pi_{ik}(\hat{w}) \hat{\beta}_k^T r_i, \quad (6)$$

where $\hat{\theta}$ is the parameter vector obtained at the convergence of the EM algorithm. Thus, since this expectation is a sum of polynomials weighted by the logistic probabilities, it is adapted for signals approximation with both smooth and abrupt transitions.

## 3.5. Case study

To perform the feature extraction, since the impedance spectrums include three regimes which correspond to three behaviors of the stack, the number of regressive components $K$ is then set to 3. The degree $p$ of the polynomial regression is set to 3, which is adapted to the different regimes in these spectrums. Figure 5 illustrates the behavior of the hidden process regression. The three domains are clearly identified.

# 4. Time duration estimation, results and discussion

The following experiments aim to illustrate the capability of the proposed approach to estimate the time duration using the features extracted from the spectrum (real and imaginary parts)..

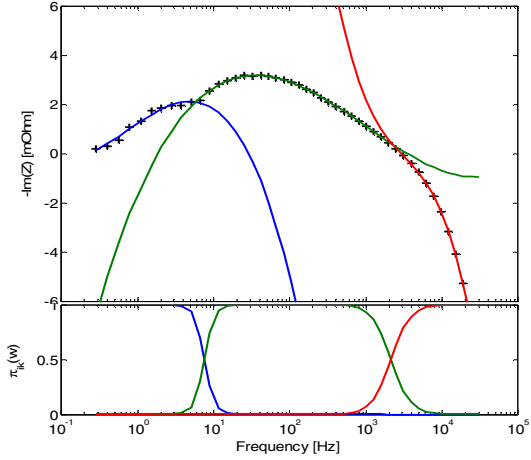

Figure 5: Original signal and the 3 polynomials (top) and their corresponding logistic probabilities (bottom) for the parametrization of imaginary part of the EIS.

To assess the performance of the proposed approach, we consider a dataset containing 29 impedance spectrum measurements carried out on two FC (FC1 and FC2 as described before) during 1000 hours. The feature extraction methods described in the previous section are applied on each spectrum measurement. These lead to extract 4 features $\{a_i\}_{1 \leq i \leq 4}$ from the real part (external logsig model) and 14 features $\{\beta_{i,j}, f_1, f_2\}_{1 \leq i,j \leq 4}$ from the imaginary part (3 hidden logistic process regressions with 12 coefficients of the three polynomial fitting of the curve and the 2 frequencies delimiting the central polynomial). So a total set of 18 descriptors are available for each impedance spectrum measurement. For the time duration estimation, we use a linear regression when the model inputs are the extracted features selected from the real part, from the imaginary part and from the combination of the two.

Considering the weak number of observations (#29), the first results obtained with the complete set of features (#18) with a cross-validation procedure, lead to very bad time duration estimation. The curse of dimensionality (cf. section 3) is clearly reached and a preliminary feature selection is highly recommended.
The selection of a subset of relevant features among the initial ones is achieved by an exhaustive search. All the possible parameter combinations are evaluated in terms of mean square error of the duration time. The chosen combination is the one which leads to the minimal rate of the test error.

Table 1 summarizes the results obtained on the whole data set (FC1 and FC2) for different model inputs. The performances were evaluated by splitting the whole data set on a training set and a test set and computing the mean error (ME) obtained on the two sets (expressed in hours). Because of the small size of the available data set, the "leave one out" method has been carried out [13]. Thus, 29 trainings have been made that correspond to 29 different training and test sets where each one of them consists to omit a sample in the training phase, on which the performance is evaluated in the test phase. The data have been normalized. The 3 rows of Table 1 correspond to the feature selection operated from the real part of the impedance spectrum descriptors, from the imaginary part and from the combination of the two. It can be noticed that the best subset extracted from the whole spectrum does not aggregate the best subset from the real part to the best subset obtained from the imaginary part.

Table 1: Linear regression model for the duration time estimation using different input descriptors.

| Mean error (in hours) | **Linear Regression** | |
|---|---|---|
| | Training set | Test set |
| Real part (dim=1) $\{a_2\}$ | 181.40 | 194.02 |
| Imag. part (dim=3) $\{\beta_{23}, \beta_{32}, \beta_{34}\}$ | 137.06 | 153.53 |
| Real + Imag. parts (dim=7) $\{\beta_{21}, \beta_{23}, \beta_{24}, a_1, a_2, a_3, a_4\}$ | 94.80 | 142.30 |

We can see that regression using features extracted from both real and imaginary parts of the impedance spectrum are better than those obtained with one kind of features. Complementary information is indeed extracted from the joint use of these two parts of the impedance spectrum and that can give a better insight of the behaviour of the fuel cell. Moreover, it can be noticed that the optimal subset of features contains three coefficients of the polynomial fitting the central part of the imaginary impedance and all the parameters of the external model approximating the real part.

Figure 6 illustrates the distribution of ME over all the duration times among the database. The left part corresponds to the training phase and the right part the test phase. With such a model, the duration time of the fuel cell can be estimated with a mean error of 142 hours over the total duration of 1000 hours. A neural network regression model has been also tested to evaluate the benefits of incorporating non linearity. Even if the training errors can decrease, the obtained results in the test phase are similar if not lower compared to those of linear regression (over-learning). This does not indicate that the problem is linear. It means that improving performance requires additional data including measurements carried out on different fuel cells in different configurations.

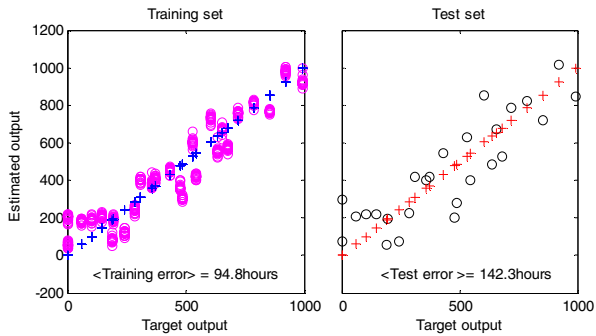

Figure 6: Duration time estimation over the training (left) and the test (right) sets for the linear regression

The curse of dimensionality is illustrated on Figure 7 that presents the target errors for all the combinations of the initial 18 features. It can be seen that errors grow in the case of large size of feature subset.

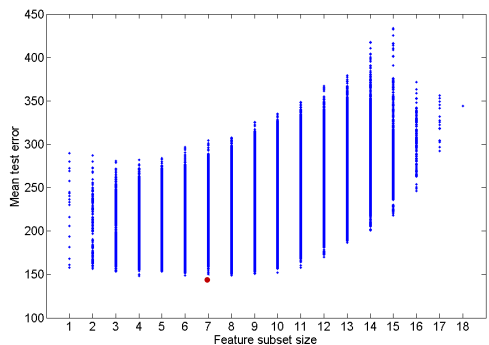

Figure 7: Evolution of the mean test error (in hours) function of the feature subset dimension

## 5. Conclusion

This article has presented a fuel cell time life estimation approach from EIS measurement. It involves a parametrization of both real and imaginary parts of the impedance, a feature selection procedure to keep only relevant descriptors and a regression model. While the parametrization of the real part simply uses an external model fitting, a specific regression model incorporating a discrete hidden logistic process has been used for the imaginary part. It allows partitioning the measurement into three parts corresponding to three behaviors of the stack, on which polynomial fittings are performed. Because of the small size of the available data set, a particular attention has been made to the feature extraction and selection steps.

Simulations on real data set have shown that FC time life can be estimated with a mean error of 142 hours over a global operating duration of 1000 hours.

Additional features extracted from another characterization measurement (polarization curve) and/or additional measurements can be helpful to decrease the error rate.

Further studies must be carried out with a more exhaustive data set. Comparison with another kind of feature extraction approach using hyper-parameters extracted from the impedance spectrum can also be carried out.